\documentclass[letterpaper]{article} 
\usepackage[]{aaai2026}  
\usepackage{times}  
\usepackage{helvet}  
\usepackage{courier}  
\usepackage[hyphens]{url}  
\usepackage{graphicx} 
\usepackage{natbib}  
\usepackage{caption} 
\usepackage{algorithm}
\usepackage{algorithmic}
\usepackage{array}
\usepackage{amsmath}
\usepackage{amsfonts}
\usepackage{booktabs} 
\usepackage{csquotes}
\usepackage{arydshln} 
\usepackage{multicol}
\usepackage{multirow}
\usepackage[table]{xcolor}   
\usepackage{enumitem}
\usepackage[most]{tcolorbox}
\newcommand{\cmark}{\ding{51}} 
\newcommand{\xmark}{\ding{55}} 
\usepackage{booktabs}
\usepackage{pifont}

\newcommand{\std}[1]{\text{\scriptsize{(#1)}}}
\newtcolorbox{prompt}[1]{
    enhanced,
    drop shadow=black!5!white,
    left=4mm,
    right=4mm,
    top=2mm,
    bottom=2mm,
    boxsep=0mm,
    rounded corners,
    title=#1,
    fontupper=\footnotesize\linespread{0.9}\fontfamily{lmr}\selectfont,
    }
\usepackage{newfloat}
\usepackage{listings}
\DeclareCaptionStyle{ruled}{labelfont=normalfont,labelsep=colon,strut=off} 
\floatstyle{ruled}
\newfloat{listing}{tb}{lst}{}
\floatname{listing}{Listing}
\title{Teaching According to Students' Aptitude: Personalized Mathematics Tutoring via Persona-, Memory-, and Forgetting-Aware LLMs}
\author{
    Yang Wu\textsuperscript{\rm 1}, Rujing Yao\textsuperscript{\rm 2},
    Tong Zhang\textsuperscript{\rm 2},
    Yufei Shi\textsuperscript{\rm 3},
    Zhuoren Jiang\textsuperscript{\rm 4},
    Zhushan Li\textsuperscript{\rm 5},
    Xiaozhong Liu\textsuperscript{\rm 1}\thanks{Corresponding author.}
}
\affiliations{
    \textsuperscript{\rm 1}Worcester Polytechnic Institute, United States\quad
    \textsuperscript{\rm 2}Nankai University, China\\
    \textsuperscript{\rm 3}The Hong Kong Polytechnic University, China \quad
    \textsuperscript{\rm 4}Zhejiang University, China\quad
    \textsuperscript{\rm 5}Boston College, United States
    
    \{ywu19, xliu14\}@wpi.edu,\quad \{rjyao, tongzhangnk\}@mail.nankai.edu.cn,\\
    yufei1999.shi@connect.polyu.hk,\quad jiangzhuoren@zju.edu.cn,\quad zhushan.li@bc.edu
}

\usepackage{bibentry}

\begin{document}

\maketitle

\begin{abstract}

Large Language Models (LLMs) are increasingly integrated into intelligent tutoring systems to provide human-like and adaptive instruction. However, most existing approaches fail to capture how students' knowledge evolves dynamically across their proficiencies, conceptual gaps, and forgetting patterns. This challenge is particularly acute in mathematics tutoring, where effective instruction requires fine-grained scaffolding precisely calibrated to each student's mastery level and cognitive retention. To address this issue, we propose TASA (\textbf{T}eaching \textbf{A}ccording to \textbf{S}tudents' \textbf{A}ptitude), a student-aware tutoring framework that integrates persona, memory, and forgetting dynamics for personalized mathematics learning. Specifically, TASA maintains a structured student persona capturing proficiency profiles and an event memory recording prior learning interactions. By incorporating a continuous forgetting curve with knowledge tracing, TASA dynamically updates each student's mastery state and generates contextually appropriate, difficulty-calibrated questions and explanations. Empirical results demonstrate that TASA achieves superior learning outcomes and more adaptive tutoring behavior compared to representative baselines, underscoring the importance of modeling temporal forgetting and learner profiles in LLM-based tutoring systems.\footnote{https://github.com/YANGWU001/TASA}

\end{abstract}


\section{Introduction}

Education is a primary engine of social and economic development, but access to high-quality educational resources remains uneven across regions and communities. Recent progress in large language models (LLMs) indicates that AI-supported tutoring can deliver high-quality guidance to substantially more learners~\citep{openai_gpt5_2025,anthropic_claude4_2024}. However, many LLM tutors follow a general-purpose design and produce uniform responses that do not adapt to an individual learner’s ongoing needs~\citep{kasneci_chatgpt_education_2023,baidoo2023education,wu2023community}. Prior personalization methods~\citep{park2024empowering,liu2025one} condition responses on a learner’s past interactions, relying on a static memory assumption that retrieves past records as context without modeling how memory and retention vary over time. This overlooks well-established cognitive reality—learners, especially children, do not acquire knowledge monotonically; they forget, revisit, and reconstruct understanding through time-dependent processes~\citep{ebbinghaus2013,cepeda_2006}. Learners also vary in learning pace and retention, whereas current storage and retrieval strategies often ignore such variability. The issue is salient in mathematics, where mastery depends on sustained practice and timely review; proficiency can degrade within weeks or months without reinforcement~\citep{chen2025effects,rohrer_taylor_2007}. Ignoring temporal dynamics and inter-learner variability reduces the realism of personalized LLM tutoring and limits its effectiveness for continuous, long-term support.

\begin{figure}[t]
\centering 
\includegraphics[width=\linewidth]{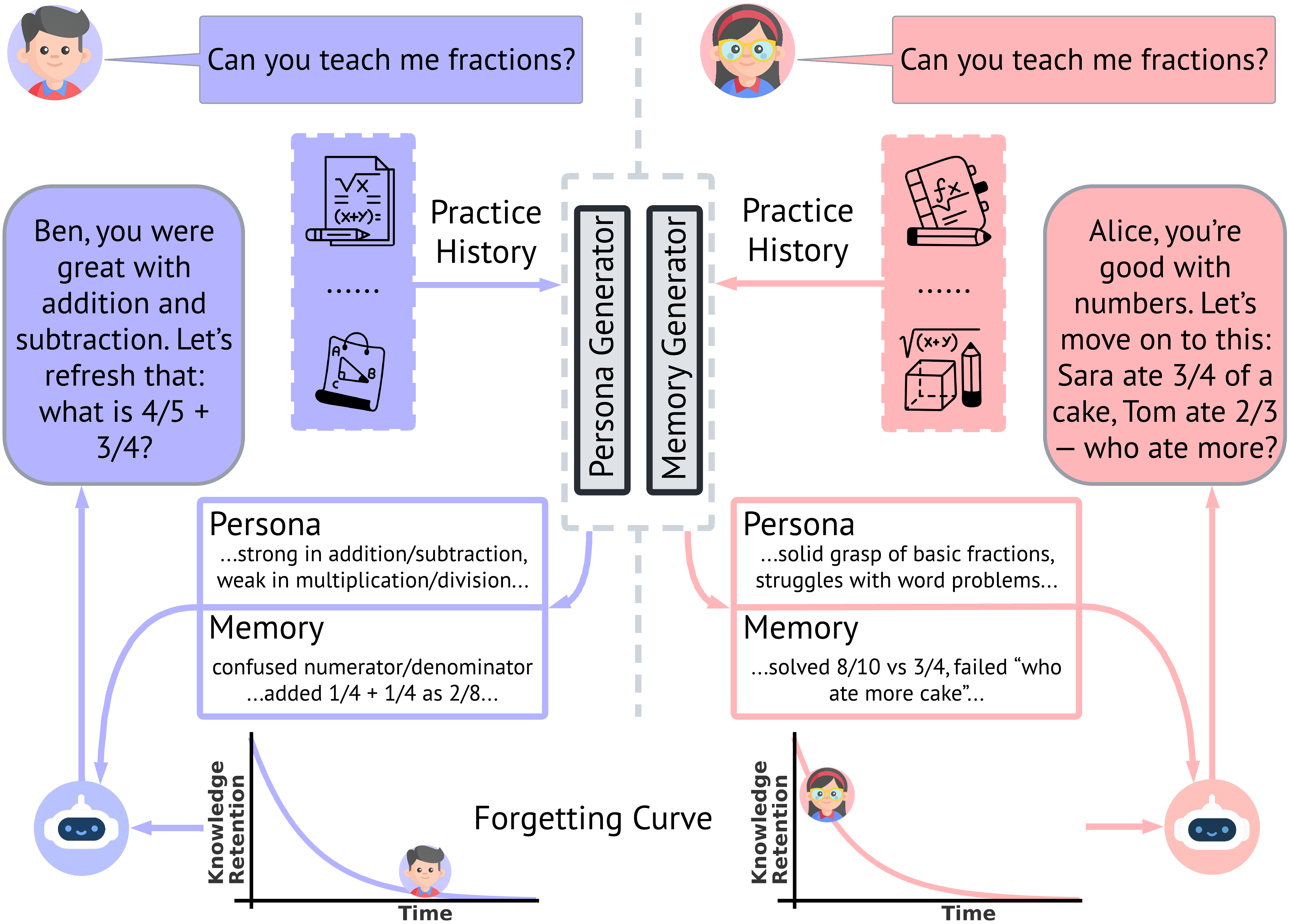}
\caption{Given the same query, TASA generates personalized tutoring responses for two students based on their forgetting-adjusted persona and memory.} 
\label{fig:toy_example}
\end{figure}

Building on these principles, we propose Teaching According to Students' Aptitude (TASA), a personalized mathematics tutoring framework that integrates three synergistic components into a unified LLM tutor: persona modeling, event memory, and forgetting-aware dynamics. As shown in Figure~\ref{fig:toy_example}, TASA constructs for each student a persona profile that summarizes proficiency and traits across skills and maintains an event memory that records recent problem-solving episodes. Crucially, unlike existing personalization methods that treat retrieved student information as static snapshots, TASA continuously estimates knowledge retention by applying a temporal decay model grounded in cognitive psychology and modulates both persona strengths and memory traces according to individualized forgetting curves. This forgetting-aware mechanism enables the tutor to generate instruction based on temporally decayed mastery estimates rather than static historical records. As illustrated in Figure~\ref{fig:toy_example}, when two students pose the same query about fractions, TASA generates distinct tutoring responses tailored to their retention states: for a student with high forgetting scores and low knowledge retention in previously mastered concepts, it prioritizes review with foundational practice; for another with low forgetting scores and stable retention, it directly addresses identified weaknesses with advanced problems. Specifically, TASA dynamically adjusts tutoring responses by applying temporal decay to retrieved persona and memory: it computes concept-level forgetting scores from knowledge tracing estimates and elapsed practice intervals, then uses these scores to rewrite student profiles before instruction generation. This temporal modeling captures the nature of learning, where mastery degrades without practice and consolidates through spaced review, thereby calibrating instruction to each student's current retention state instead of relying solely on historical performance.

Our contributions in this paper are threefold and can be summarized as follows:

$\bullet$ We present TASA, an LLM-driven mathematics tutoring framework that jointly models each student's persona, event memory, and forgetting dynamics, enabling personalized mathematics instruction.\par 
$\bullet$ We formalize personalized mastery decay with a learnable forgetting curve, grounded in cognitive psychology and knowledge tracing, to estimate concept-level temporal mastery decay and to modulate both persona strengths and memory traces within the tutoring loop. \par 
$\bullet$ We conduct empirical studies in mathematics tutoring scenarios and show that TASA outperforms state-of-the-art LLM tutors in learning gains, learning efficiency, and personalization quality.\par

\section{Related Work}

\paragraph{Large Language Models for Education}
Large language models (LLMs) have been widely applied to various educational tasks, including automated feedback~\citep{guo2024using,nair2024closing,zhang2025sefl,du2024llms}, learning resource recommendation~\citep{zaiane2002building,li2024learning,shahzad2025comprehensive,abu2024knowledge,chu2025llmagentseducationadvances}, and conversational teaching assistance~\citep{hu2024foke,mithun2025ai,zha2024designing,wu2024rose}. Mathematics education has emerged as a particularly active domain for LLM-based solutions~\citep{mitra2024orca,swan2023math,yan2024survey}. \citet{yan2025mathagent} present MathAgent, a Mixture-of-Math-Agent framework for K--12 scenarios that performs multimodal error detection and converts visual inputs into formal expressions. \citet{wu2023mathchat} introduce MathChat, which frames mathematical problem solving as a dialogue between a chat-oriented LLM agent and a tool-using user agent, enabling iterative reasoning with tool calls. \citet{lei2024macm} analyze the limitations of single LLMs on complex mathematical logical deduction and propose a multi-agent mechanism that partitions reasoning roles to improve solution quality. Related efforts span other domains as well, including physics, chemistry, and biology~\citep{ma2024llm,m2024augmenting,huang2024protchat}. In contrast to these general deployments, our work focuses on a personalized mathematics tutor that incorporates student-specific dynamics to guide practice and review.

\paragraph{Personalized Large Language Model Tutoring}
Personalized LLM tutoring adapts instruction to individual learners by modeling cognitive, affective, and preference-based dimensions. Existing methods coordinate interaction, reflection, and reaction components through memory modules and meta-agents~\citep{chen2024empowering,wang2025learnmate}, maintaining structured student profiles that encode knowledge proficiency, comprehension levels, and misconceptions~\citep{liu2024personalized,liu2024personality}. These profiles enable tutors to tailor explanations, adjust difficulty, and select content dynamically based on learner state. Extending beyond cognitive factors, recent work incorporates affective dimensions such as motivation and self-efficacy to inform feedback tone and pacing~\citep{park2024empowering}, alongside personality traits that guide instructional modality~\citep{wang2025llm,roccas2002big,sonlu2024effects,wu2024knowledge}. In mathematics education, these principles manifest as conversation-based tutoring with diagnostic assessments~\citep{chudziak2025ai,yao2025elevating}, learning style modeling integrated with Socratic dialogue~\citep{liu2025one}, and stepwise problem-solving with adaptive feedback~\citep{hsu2025mathedu,wu2025active}. While these approaches effectively personalize content by leveraging learner profiles and interaction histories, they treat retrieved student information as static snapshots. In contrast, our work dynamically adjusts persona and memory through individualized forgetting curves, aligning tutoring behavior with the natural temporal dynamics of human learning and retention.

\paragraph{Knowledge Tracing}
Knowledge tracing (KT) aims to monitor and predict students' evolving mastery states by modeling their interactions with learning materials over time. Traditional approaches such as BKT~\citep{corbett1994knowledge} and DKT~\citep{piech2015deep} use statistical or neural models to estimate proficiency and predict future performance. Recent work has enhanced these models by incorporating temporal dynamics. MemoryKT~\citep{lin2025memorykt} uses a temporal variational autoencoder to capture knowledge evolution, while LefoKT~\citep{bai2025rethinking} introduces a unified attention-based architecture that explicitly models relative forgetting patterns. Beyond predictive modeling, several studies leverage LLM agents for more interactive and personalized knowledge tracing. \citet{yang2024content} design a multi-agent system where administrator, judger, and critic agents collaborate to assess student cognitive states. EduAgent~\citep{xu2024eduagent} simulates students as diverse personas to adaptively trace knowledge progression, and dialogue-driven approaches~\citep{scarlatos2025exploring,yao2025intelligent} dynamically refine mastery estimation through conversational exchanges. TutorLLM~\citep{li2025tutorllm} combines knowledge tracing with RAG to retrieve relevant historical interactions for guiding content generation. While these KT methods primarily estimate mastery for prediction, our approach leverages forgetting dynamics to directly modulate tutoring content generation, bridging knowledge tracing with temporally-aware instructional adaptation.

\section{Methodology}
\label{sec:methodology}

\begin{figure*}[t]
  \centering
  \includegraphics[width=1.0\linewidth]
  {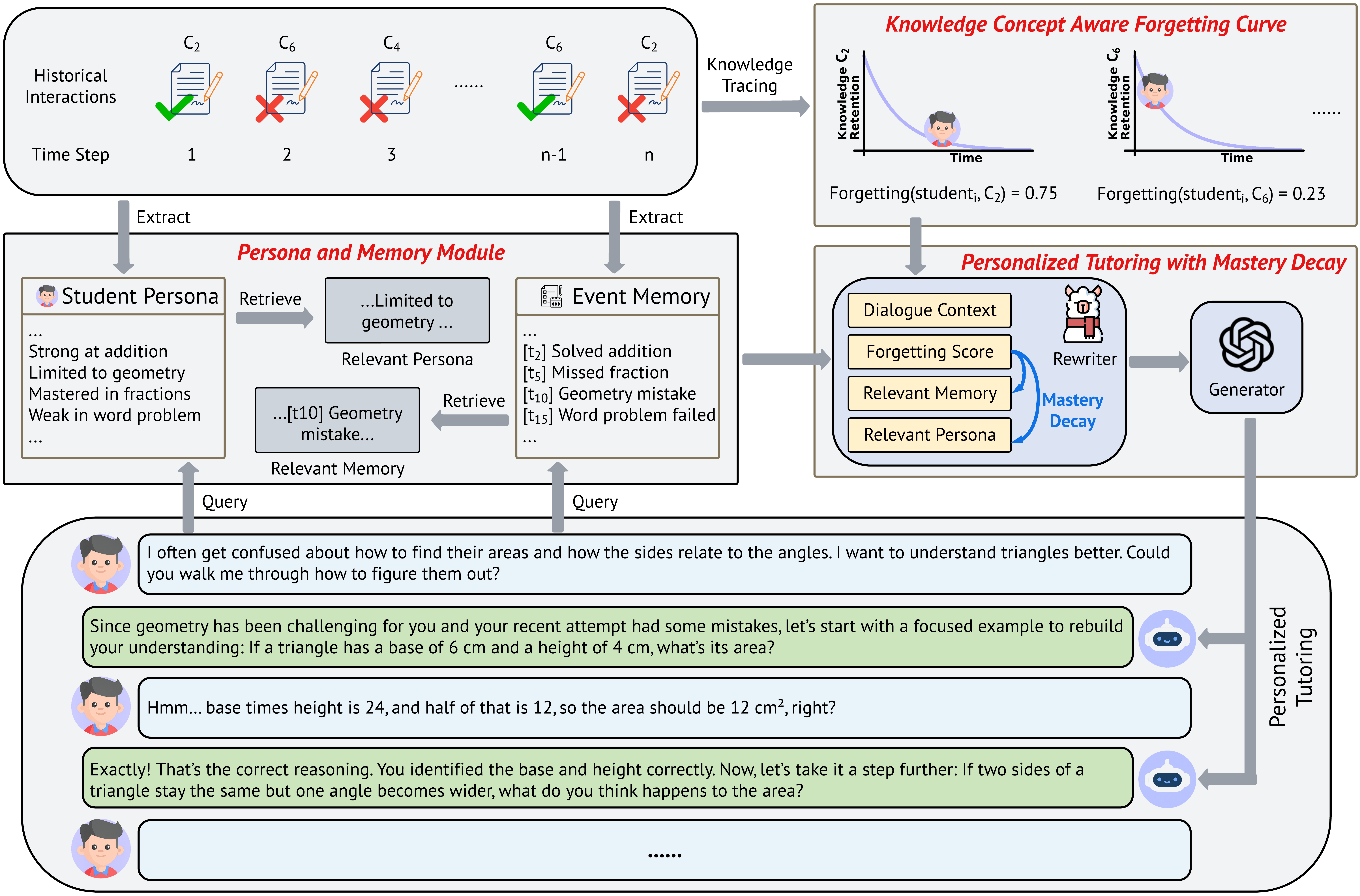}
  \caption{Illustration of our proposed TASA framework. TASA first extracts the student’s persona and event memory from historical interactions. During tutoring, it retrieves relevant persona and memory, applies forgetting-aware decay through a Rewriter, and generates personalized instruction via a Generator conditioned on the decayed student state.}
  \label{fig:framework}
\end{figure*}

In this section, we propose Teaching According to Students’ Aptitude (TASA), a novel personalized mathematics tutoring framework that dynamically integrates students' persona and memory with forgetting patterns captured from their learning trajectory. We first introduce the task definition of personalized mathematics tutoring, and then present TASA in detail with the overall framework illustrated in Figure~\ref{fig:framework}.

\subsection{Preliminary Definitions}
The goal of personalized mathematics tutoring is to generate appropriate instructional content $z_n$ (e.g., adaptive practice questions and tailored explanations) by leveraging both the student's current tutoring dialogue session $\mathcal{H}$ and their historical interaction records $\mathcal{X}$. Formally, the current tutoring session is represented as: 
$\mathcal{H} = \{(h_1, z_1), (h_2, z_2), \ldots, (h_{n-1}, z_{n-1}), h_n\}$, where each $h_i$ denotes the student's $i$-th query and $z_i$ denotes the corresponding personalized instructional content, with $n$ being the number of interaction turns in the current session. The student's prior learning trajectory is defined as a sequence of question--response interactions. Each historical sequence at time step $t$, denoted as $\mathcal{X}_t = \langle x_1, x_2, \ldots, x_t \rangle$,
consists of interaction tuples $x_i = (q_i, \{c\}, r_i, t_i)$,
where $q_i$ is the question presented, $\{c\} \subseteq \mathcal{C}$ represents the set of knowledge concepts (KCs) associated with $q_i$ from a predefined skill taxonomy $\mathcal{C}$, $r_i \in \{0,1\}$ indicates the correctness of the student's response, and $t_i$ records the timestamp of the interaction. Unlike conventional personalized tutoring systems that treat historical records as static retrieval contexts, our method dynamically models temporal knowledge decay to generate pedagogically adaptive and contextually coherent instruction $z_n$ that reflects each student's evolving cognitive state and individualized retention patterns.

\subsection{Approach Overview}
TASA first constructs each student's persona representation, event memory bank, and individualized forgetting curve from their historical learning trajectory $\mathcal{X}$. During tutoring sessions, given the student's current query $h_n$ in $\mathcal{H}$, TASA retrieves relevant persona traits and memory entries, estimates forgetting scores for the query-related knowledge concepts to decay the retrieved information according to temporal retention dynamics, and finally generates personalized instructional content $z_n$ via an LLM conditioned on the forgetting-adjusted student state.

\subsection{Student Modeling with Persona and Memory}

Each student is represented by two complementary modules that capture learning patterns at different temporal granularities: persona and event memory. The persona encodes stable proficiency traits across knowledge concepts, while event memory records fine-grained historical problem-solving experiences. We describe their extraction, storage, and retrieval mechanisms below.

\subsubsection{Extraction}
Given a student's historical interaction sequence $\mathcal{X}_t$, we employ two specialized agent-based generators, \textit{Persona Generator} and \textit{Memory Generator}, to excavate and summarize learning patterns using the zero-shot reasoning capabilities of LLMs. The Persona Generator analyzes the full trajectory $\mathcal{X}_t$ to produce a structured profile summarizing the student's mastery level across knowledge concepts $\mathcal{C}$. Each persona entry $\text{p}_j$ is represented as:
\begin{equation}
\text{p}_j = (d_j^{\text{persona}}, \{c\}_j^{\text{persona}}),
    \label{eq:persona_generator}
\end{equation}
where $d_j^{\text{persona}}$ is a natural-language description (e.g., \enquote{Student excels at basic arithmetic but struggles with multi-step word problems}), and $\{c\}_j^{\text{persona}} \subseteq \mathcal{C}$ denotes the associated knowledge concepts extracted as keywords. The Memory Generator processes recent interactions to construct event-level records. Each memory entry $\text{m}_j$ captures a specific learning episode:
\begin{equation}
\text{m}_j = (t_j, d_j^{\text{mem}}, \{c\}_j^{\text{mem}}),
    \label{eq:memory_generator}
\end{equation}
where $t_j$ is the timestamp, $d_j^{\text{mem}}$ is a brief summary of the event (e.g., \enquote{Incorrectly added $1/4 + 1/4$ as $2/8$ in Q15}), and $\{c\}_j^{\text{mem}}$ are the relevant KCs as keywords.

\subsubsection{Storage} 
The extracted persona and memory entries are encoded into dense vector representations and stored in two separate banks: $\mathcal{M}^{\text{persona}}$ for persona traits and $\mathcal{M}^{\text{mem}}$ for event records. Formally:
\begin{equation}
\resizebox{\dimexpr\columnwidth-2.2em\relax}{!}{$
\begin{aligned}
\mathcal{M}^{\text{persona}}
&= \{ \phi(d_j^{\text{persona}}), \phi(\{c\}_j^{\text{persona}}) \mid j \in \{1,\ldots,L_p\} \},\\
\mathcal{M}^{\text{mem}}
&= \{ (t_j, \phi(d_j^{\text{mem}}), \phi(\{c\}_j^{\text{mem}})) \mid j \in \{1,\ldots,L_m\} \}
\end{aligned}
$}
\end{equation}
where $L_p$ and $L_m$ denote the total number of persona and memory entries respectively, and $\phi(\cdot)$ denotes text encoder (e.g., BGE encoder~\citep{bge_embedding}) that maps textual descriptions and keyword sets into a shared embedding space. Each entry thus contains both semantic content (description embedding) and structured metadata (KC keyword embeddings), enabling effective information retrieval.

\subsubsection{Retrieval}
Within a tutoring session $\mathcal{H}$, given the student's current query $h_n$, we retrieve the most relevant persona traits and memory events from $\mathcal{M}^{\text{persona}}$ and $\mathcal{M}^{\text{mem}}$ respectively through a two-stage process. We first compute a hybrid similarity score $\mathcal{S}$ for each entry:
\begin{equation}
\begin{aligned}
\mathcal{S}_j^{\text{persona}}(h_n, \text{p}_j)
&= \lambda \cos\bigl(\phi(h_n), \phi(d_j^{\text{persona}})\bigr) +\\
&\quad (1-\lambda) \cos\bigl(\phi(h_n), \phi(\{c\}_j^{\text{persona}})\bigr),\\[2pt]
\mathcal{S}_j^{\text{mem}}(h_n, \text{m}_j)
&= \lambda \cos\bigl(\phi(h_n), \phi(d_j^{\text{mem}})\bigr) +\\
&\quad (1-\lambda) \cos\bigl(\phi(h_n), \phi(\{c\}_j^{\text{mem}})\bigr).
\end{aligned}
\label{eq:lambda}
\end{equation}
where the first term measures semantic similarity between query and description, the second term captures concept-level alignment via keywords, and $\lambda$ is a weight that balances the two terms (default $\lambda = 0.5$). We select the top-$K$ entries from each bank based on the highest scores. To refine the selection, we apply a cross-encoder reranker (e.g., BGE reranker~\citep{li2023making}) and retain the top-3 from each bank:
\begin{equation}
\begin{aligned}
\{\text{p}_1^*, \text{p}_2^*, \text{p}_3^*\} &= \text{Rerank}_{\text{top-3}}\bigl(\{\text{p}_j\}_{j=1}^K, h_n\bigr), \\
\{\text{m}_1^*, \text{m}_2^*, \text{m}_3^*\} &= \text{Rerank}_{\text{top-3}}\bigl(\{\text{m}_j\}_{j=1}^K, h_n\bigr).
\end{aligned}
\label{eq:rerank}
\end{equation}
These retrieved entries are then used in subsequent tutoring generation to provide student-specific context.
\subsection{Forgetting Curve over Knowledge Concepts}

\subsubsection{Knowledge Tracing}
To enable temporally adaptive tutoring, TASA introduces a forgetting score $F_c(t)$ that quantifies the likelihood of a student forgetting concept $c$ at time $t$. The score ranges from $0$ to $1$, where higher values indicate higher forgetting probability. It is designed to be continuous, monotonically increasing with respect to elapsed time $\Delta t_c$ since the last practice of concept $c$, and monotonically decreasing with respect to the current mastery probability $s_{t,c}$ estimated by a knowledge tracing (KT) model. Formally, the mastery probability is computed as:
\begin{equation}
    s_{t,c} = \mathcal{G}(\mathcal{X}_t, c),
\end{equation}
where $\mathcal{G}(\cdot)$ denotes the KT model that predicts mastery given the student's interaction history $\mathcal{X}_t$ and concept $c$. In our implementation, we use DKT~\citep{piech2015deep} as the instantiation of $\mathcal{G}$.

\subsubsection{Forgetting Score Definition}
Following cognitive theories of exponential memory decay~\citep{rubin1996one,wixted1991form}, we define:
\begin{equation}
    F_c(t) = 1 - s_{t,c} \exp\left(-\frac{\Delta t_c}{S_c}\right),
    \label{eq:forgetting}
\end{equation}
where $S_c$ denotes the student's memory strength for concept $c$. When $\Delta t_c$ increases, the exponential term decays, producing a higher forgetting score; when the mastery $s_{t,c}$ is large, the term suppresses $F_c(t)$, reflecting stronger retention. For computational efficiency, we use a smooth rational approximation:
\begin{equation}
    F_c(t) \approx (1 - s_{t,c}) \cdot \frac{\Delta t_c}{\Delta t_c + \tau},
    \label{eq:forgetting_approx}
\end{equation}
where $\tau$ is a stability constant ensuring numerical continuity for small time intervals. Detailed derivation is provided in Appendix B.

\subsection{Personalized Tutoring Generation}

Given the retrieved persona traits and memory events, TASA generates the next instructional content \(z_n\) that both provides feedback on the student’s most recent answer and proposes the next practice question calibrated to the current retention state. The process first rewrites retrieved evidence with forgetting score and then generates \(z_n\) conditioned on the adjusted student state.

\subsubsection{Forgetting-Aware Rewriting}

The retrieved persona \(\{\text{p}_1^*, \text{p}_2^*, \text{p}_3^*\}\) and memory \(\{\text{m}_1^*, \text{m}_2^*, \text{m}_3^*\}\) summarize historical proficiency and learning episodes, yet concept-level mastery evolves over time due to forgetting. For every retrieved description that references a concept \(c\), we compute the forgetting score \(F_c(t)\) as in Equation~\ref{eq:forgetting_approx} and revise the mastery to reflect this time-dependent decay. We adopt a tuning-free procedure that uses the zero-shot capability of a large language model to decay mastery in relevant persona and memory descriptions given the prompt:

\begin{tcolorbox}[
  colframe=black!50,
  arc=0mm,
  boxrule=0.5pt,
  width=\linewidth,
  left=1.5mm,
  right=1.5mm,
  top=1mm,
  bottom=1mm,
  breakable
]
{\small\ttfamily
[System Message]\par
You are a personalized math tutor. Given a student's original state for a concept, including mastery, last practice interval, and forgetting score, rewrite the description to reflect time-dependent forgetting. Output only the updated description, concise and specific to the concept.\par
\par
[User Message]\par
The student's original state: ``\textit{[description $d_j$]}'' for concept \textit{[c]}, with mastery \textit{[$s_{t,c}$]}.\par
This concept was last practiced \textit{[$\Delta t_c$]} days ago.\par
Current forgetting score: \textit{[$F_c(t)$]}.
}
\end{tcolorbox}\par\noindent
Applying this rewrite to the top retrieved items yields forgetting-adjusted persona and memory
\[
\{\tilde{\text{p}}_1, \tilde{\text{p}}_2, \tilde{\text{p}}_3\}, \quad \{\tilde{\text{m}}_1, \tilde{\text{m}}_2,\tilde{\text{m}}_3\}.
\]

\subsubsection{Generating Explanatory Feedback and the Next Question}

Conditioned on \(\mathcal{H}\), the adjusted persona \(\{\tilde{\text{p}}_j\}_{j=1}^{3}\), and the adjusted memory \(\{\tilde{\text{m}}_j\}_{j=1}^{3}\), TASA produces \(z_n\) that (a) delivers a concise explanation addressing the student’s latest answer and (b) presents the next practice question tailored to the current retention estimate.\footnote{The details of the personalized tutoring prompt are provided in Appendix C.} Formally, we define the generation process as:
\begin{equation}
z_n = \text{Generator}\!\left(\mathcal{H}, \{\tilde{\text{p}}_j\}_{j=1}^{3}, \{\tilde{\text{m}}_j\}_{j=1}^{3}\right).
    \label{eq:generation_z}
\end{equation}
After generation, \(z_n\) is appended to the current session:
\begin{equation}
\mathcal{H} \leftarrow \mathcal{H} \cup z_n.
    \label{eq:update_h}
\end{equation}
When the session ends, the accumulated interactions are merged into the long-term trajectory for subsequent retrieval and forgetting updates:
\begin{equation}
\mathcal{X} \leftarrow \mathcal{X} \cup \mathcal{H}.
    \label{eq:update_x}
\end{equation}

\section{Experiments}

\begin{table*}[t]
    \centering
    \renewcommand{\arraystretch}{1.2}
    \resizebox{1.0\textwidth}{!}{
    \begin{tabular}{@{}llcccccccc@{}}
        \toprule
        \multirow{2}{*}[-1ex]{LLM Backbone} 
        & \multirow{2}{*}[-1ex]{Method} 
        & \multicolumn{4}{c}{Normalized Learning Gain ($\Delta$-NLG \%)} 
        & \multicolumn{4}{c}{Response Personalization (Win Rate \%)} \\
        \cmidrule(lr){3-6} \cmidrule(lr){7-10}
        & & Assist2017 & NIPS34 & Algebra2005 & Bridge2006 
          & Assist2017 & NIPS34 & Algebra2005 & Bridge2006 \\
        \midrule
        \multirow{5}{*}{gpt-oss-120b} 
            & Vanilla-ICL         & \underline{44.8~\std{1.4}} & 35.2~\std{0.8} & 46.1~\std{2.1} & 36.0~\std{1.2} & - & - & - & - \\
            & MathChat            & 23.0~\std{1.9} & 34.9~\std{1.3} & 36.8~\std{1.0} & 23.7~\std{1.8} & 21.4~\std{2.2} & 7.7~\std{0.9} & \underline{68.8~\std{1.5}} & 15.4~\std{2.3} \\
            & TutorLLM & 43.6~\std{1.1} & 25.1~\std{1.4} & \underline{62.1~\std{2.0}} & \underline{40.3~\std{1.7}} & \underline{65.7~\std{1.6}} & \underline{67.1~\std{1.9}} & 55.0~\std{1.3} & \underline{50.0~\std{2.2}} \\
            & PSS-MV              & 23.6~\std{1.3} & \underline{43.5~\std{1.8}} & 50.5~\std{1.2} & 31.1~\std{1.0} & 52.7~\std{0.9} & 58.0~\std{1.4} & 60.0~\std{2.4} & 38.8~\std{1.3} \\
            & \textbf{TASA (Ours)} & \textbf{54.1~\std{1.7}} & \textbf{45.6~\std{1.9}} & \textbf{67.4~\std{2.3}} & \textbf{44.5~\std{1.5}} & \textbf{86.4~\std{1.8}} & \textbf{79.4~\std{2.1}} & \textbf{82.5~\std{1.0}} & \textbf{75.6~\std{2.3}} \\
        \midrule
        \multirow{5}{*}{Qwen3-4B-Instruct} 
            & Vanilla-ICL         & 47.6~\std{1.3} & 36.5~\std{1.0} & 61.9~\std{2.0} & 35.4~\std{1.2} & - & - & - & - \\
            & MathChat            & 25.5~\std{1.9} & 22.2~\std{1.6} & 41.7~\std{1.1} & 36.6~\std{1.5} & 47.1~\std{1.9} & 35.3~\std{2.3} & 37.5~\std{1.6} & 58.8~\std{2.4} \\
            & TutorLLM & 38.0~\std{2.0} & 16.3~\std{1.4} & 51.8~\std{1.5} & \underline{39.5~\std{1.8}} & \textbf{88.2~\std{1.3}} & \underline{47.1~\std{2.1}} & \underline{75.0~\std{1.8}} & \underline{73.3~\std{1.7}} \\
            & PSS-MV              & \underline{53.0~\std{1.5}} & \underline{38.1~\std{2.4}} & \textbf{68.4~\std{1.2}} & 37.9~\std{1.3} & 35.3~\std{1.2} & 5.9~\std{2.0} & 31.2~\std{1.8} & 23.5~\std{2.3} \\
            & \textbf{TASA (Ours)} & \textbf{56.7~\std{2.0}} & \textbf{40.3~\std{1.6}} & \underline{66.7~\std{1.9}} & \textbf{42.8~\std{2.1}} & \underline{76.8~\std{1.7}} & \textbf{62.3~\std{2.2}} & \textbf{82.4~\std{1.6}} & \textbf{88.2~\std{1.9}} \\
        \midrule
        \multirow{5}{*}{Llama3.1-8B-Instruct} 
            & Vanilla-ICL         & 45.2~\std{1.4} & 33.4~\std{1.2} & 58.1~\std{1.5} & 35.5~\std{1.9} & - & - & - & - \\
            & MathChat            & 33.9~\std{2.1} & 31.1~\std{1.8} & 56.6~\std{1.4} & 31.8~\std{1.7} & 52.9~\std{1.8} & 64.7~\std{1.9} & 50.0~\std{2.3} & 70.6~\std{1.4} \\
            & TutorLLM & \underline{45.8~\std{1.7}} & \underline{50.1~\std{1.6}} & 62.1~\std{2.2} & 36.5~\std{1.8} & 11.1~\std{2.1} & \underline{76.5~\std{2.0}} & 41.7~\std{1.5} & 82.4~\std{1.7} \\
            & PSS-MV              & 41.6~\std{1.3} & 26.3~\std{1.5} & \textbf{64.9~\std{1.1}} & \underline{43.9~\std{1.9}} & \underline{76.5~\std{1.2}} & 54.5~\std{2.3} & \underline{56.2~\std{1.8}} & \underline{88.2~\std{1.6}} \\
            & \textbf{TASA (Ours)} & \textbf{59.4~\std{2.1}} & \textbf{54.9~\std{1.8}} & \underline{62.3~\std{2.4}} & \textbf{53.9~\std{2.0}} & \textbf{86.1~\std{1.5}} & \textbf{82.5~\std{1.9}} & \textbf{76.9~\std{1.4}} & \textbf{92.1~\std{2.2}} \\
        \bottomrule
    \end{tabular}
    }
    \caption{Comparative results on two tasks (Normalized Learning Gain and Response Personalization) across four benchmarks. The mean (std) over three random runs is reported. The best results are shown in \textbf{bold}, and the second-best are \underline{underlined}.}

    \label{tab:main_results}
\end{table*}

In this section, we conduct extensive experiments on four public math benchmarks to evaluate the performance of our TASA approach along with four state-of-the-art competitors. We first describe the benchmarks, the competitors, and our experimental settings. Then we report and analyze the experimental results to present the effectiveness of TASA.

\subsection{Benchmarks}
\label{sec:benchmark}

We evaluate our framework on four publicly available large-scale mathematics learning benchmarks commonly used in knowledge tracing and intelligent tutoring research: 1) ASSISTments2017~\citep{ASSISTments2017Dataset}, originating from the 2017 ASSISTments data mining competition, consisting of middle school students' responses to mathematics questions; 2) NIPS34~\citep{wang2020instructions}, drawn from Tasks 3 \& 4 at the NeurIPS 2020 Education Challenge, containing students' answers to multiple-choice diagnostic math questions collected from the Eedi platform; 3) Algebra2005~\citep{Stamper2010KDDCupDev}, sourced from the KDD Cup 2010 EDM Challenge, containing 13-14 year old students' step-level responses to Algebra questions; and 4) Bridge2006~\citep{Stamper2010KDDCupDev}, also derived from the KDD Cup 2010 EDM Challenge with similar question construction as Algebra2005. All datasets record extensive student-question interaction histories, including question IDs, corresponding knowledge components (KCs), timestamps, and correctness labels. Detailed data statistics are provided in Appendix A.

\subsection{Competitors}
\label{sec:baselines}

In our experiments, we compare our proposed TASA approach against four recent, state-of-the-art intelligent tutoring systems: 1) Vanilla-ICL \citep{Leong2024Putting}, which employs in-context learning with generative AI to personalize educational content by incorporating contextual information into prompts; 2) MathChat \citep{wu2023mathchat}, which frames mathematical problem solving as a dialogue between a chat-oriented LLM agent and a tool-using user agent for iterative reasoning; 3) TutorLLM \citep{li2025tutorllm}, which combines knowledge tracing with retrieval-augmented generation (RAG) to provide personalized learning recommendations based on students' historical interactions; and 4) PSS-MV~\citep{Liu2024PersonalityAware}, which simulates student personalities and validates tutoring responses across multiple aspects to deliver personality-aware conversational instruction. All competitors use the same students' learning history as TASA to ensure comparability.

\subsection{Experimental Settings}
\subsubsection{Implementation Details}
We use GPT-4o-mini~\citep{openai2024gpt4omini} conditioned on each student’s persona and learning history to simulate the student role, constraining the agent to exhibit authentic problem-solving behaviors while leveraging the model’s strong conversational ability. The tutoring interaction consists of 20 default dialogue turns, comprising 10 student queries and 10 corresponding tutor responses. We implement the tutoring component with several large language model backbones, including gpt-oss-120b~\citep{agarwal2025gpt}, Qwen3-4B-Instruct~\citep{yang2025qwen3}, Llama3.1-8B-Instruct~\citep{dubey2024llama}. For the knowledge tracing (KT) module, we adopt the open-source implementation of DKT~\citep{piech2015deep}, as it explicitly models students’ temporal learning trajectories. To ensure reproducibility and robustness, we report the mean and standard deviation across 5 random seeds.

\subsubsection{Tasks and Metrics}
\label{sec:metrics}
We follow the evaluation protocols of prior tutoring agent studies \citep{wang2025training,liu2025one} and assess our approach on two tasks. \textit{1) Normalized Learning Gain.}
We measure the improvement in student performance after a tutoring session using the normalized learning gain:
\begin{equation}
\Delta\mathrm{-NLG} \;=\; \frac{\mathrm{ACC}_{\mathrm{Post}} - \mathrm{ACC}_{\mathrm{Pre}}}{1 - \mathrm{ACC}_{\mathrm{Pre}}}.
    \label{eq:nlg}
\end{equation}
where $\mathrm{ACC}_{\mathrm{Pre}}$ and $\mathrm{ACC}_{\mathrm{Post}}$ are computed on a knowledge-concept–aware question pool before and after tutoring, respectively. \textit{2) Response Personalization.}
To assess the personalization of tutoring content, we employ GPT-5~\citep{openai_gpt5_2025} as an impartial judge following established LLM-as-judge protocols~\citep{zheng2023judging}. For each generated tutoring response, the judge compares the target method with the Vanilla-ICL baseline given the same student's learning state and history, and we report the Win Rate (\%) indicating how often the target method is preferred.

\begin{table*}[t]
    \centering
    \renewcommand{\arraystretch}{1.2}
    \resizebox{1.0\textwidth}{!}{
    \begin{tabular}{@{}llcccccccc@{}}
        \toprule
        \multirow{2}{*}[-1ex]{LLM Backbone} 
        & \multirow{2}{*}[-1ex]{Method} 
        & \multicolumn{4}{c}{Normalized Learning Gain ($\Delta$-NLG \%)} 
        & \multicolumn{4}{c}{Response Personalization (Win Rate \%)} \\
        \cmidrule(lr){3-6} \cmidrule(lr){7-10}
        & & Assist2017 & NIPS34 & Algebra2005 & Bridge2006 
          & Assist2017 & NIPS34 & Algebra2005 & Bridge2006 \\
        \midrule
        \multirow{5}{*}{Llama3.1-8B-Instruct} 
            & Vanilla-ICL   & 45.2~\std{1.4} & 33.4~\std{1.2} & 58.1~\std{1.5} & 35.5~\std{1.9} & - & - & - & - \\
            & w/o Memory & \underline{55.0~\std{1.6}} & 46.5~\std{1.7} & 57.3~\std{1.5} & 49.5~\std{2.0} & 64.2~\std{1.3} & 76.5~\std{2.1} & 59.1~\std{1.7} & 84.5~\std{1.6} \\
            & w/o Persona    & 47.9~\std{1.5} & 48.4~\std{1.9} & \underline{59.7~\std{1.8}} & 42.9~\std{1.6} & \underline{78.3~\std{2.3}} & 71.8~\std{1.7} & \underline{72.7~\std{1.8}} & 79.4~\std{2.2} \\
            & w/o Forgetting & 50.8~\std{1.3} & \underline{51.3~\std{1.8}} & 54.7~\std{2.2} & \underline{49.7~\std{1.5}} & 74.6~\std{1.9} & \underline{76.9~\std{1.4}} & 68.4~\std{1.6} & \underline{86.2~\std{1.9}} \\
            & \textbf{TASA (Ours)} & \textbf{59.4~\std{2.1}} & \textbf{54.9~\std{1.8}} & \textbf{62.3~\std{2.4}} & \textbf{53.9~\std{2.0}} & \textbf{86.1~\std{1.5}} & \textbf{82.5~\std{1.9}} & \textbf{76.9~\std{1.4}} & \textbf{92.1~\std{2.2}} \\
        \bottomrule
    \end{tabular}
    }
    \caption{Ablation study results on the effect of different TASA modules.}

    \label{tab:ablation_module}
\end{table*}

\subsection{Main Results}
\label{sec:main}


Table~\ref{tab:main_results} reports comparative results on Normalized Learning Gain ($\Delta\mathrm{-NLG}$) and Response Personalization (Win Rate) across four mathematics tutoring benchmarks and three LLM backbones. TASA consistently achieves the best or second-best performance in all settings, demonstrating strong generalization and robustness across architectures. Compared with the strongest baseline TutorLLM, TASA improves the average $\Delta\mathrm{-NLG}$ by 11.4\% and the Win Rate by 19.7\%, showing the benefit of jointly modeling persona, memory, and forgetting dynamics. Notably, on NIPS34 and Bridge2006, TASA achieves up to 9\% higher learning gain, underscoring the value of integrating time-dependent forgetting. Across different backbones, TASA exhibits stable gains, particularly on Llama3.1-8B-Instruct, where it surpasses TutorLLM by 8.5\% in $\Delta\mathrm{-NLG}$ and 9.9\% in personalization—highlighting its cross-architecture adaptability. 
Further qualitative evaluations with GPT-5-as-judge confirm that TASA’s responses are more contextually aligned, pedagogically consistent, and adaptive to individual learner states, reflecting improved personalization fidelity. Overall, these results validate that incorporating temporal forgetting and learner profiles enables more effective, human-like instructional adaptation.

\subsection{Ablation Study}
\label{sec:ablation}
\begin{figure}[t]
\centering
\includegraphics[width=\linewidth]{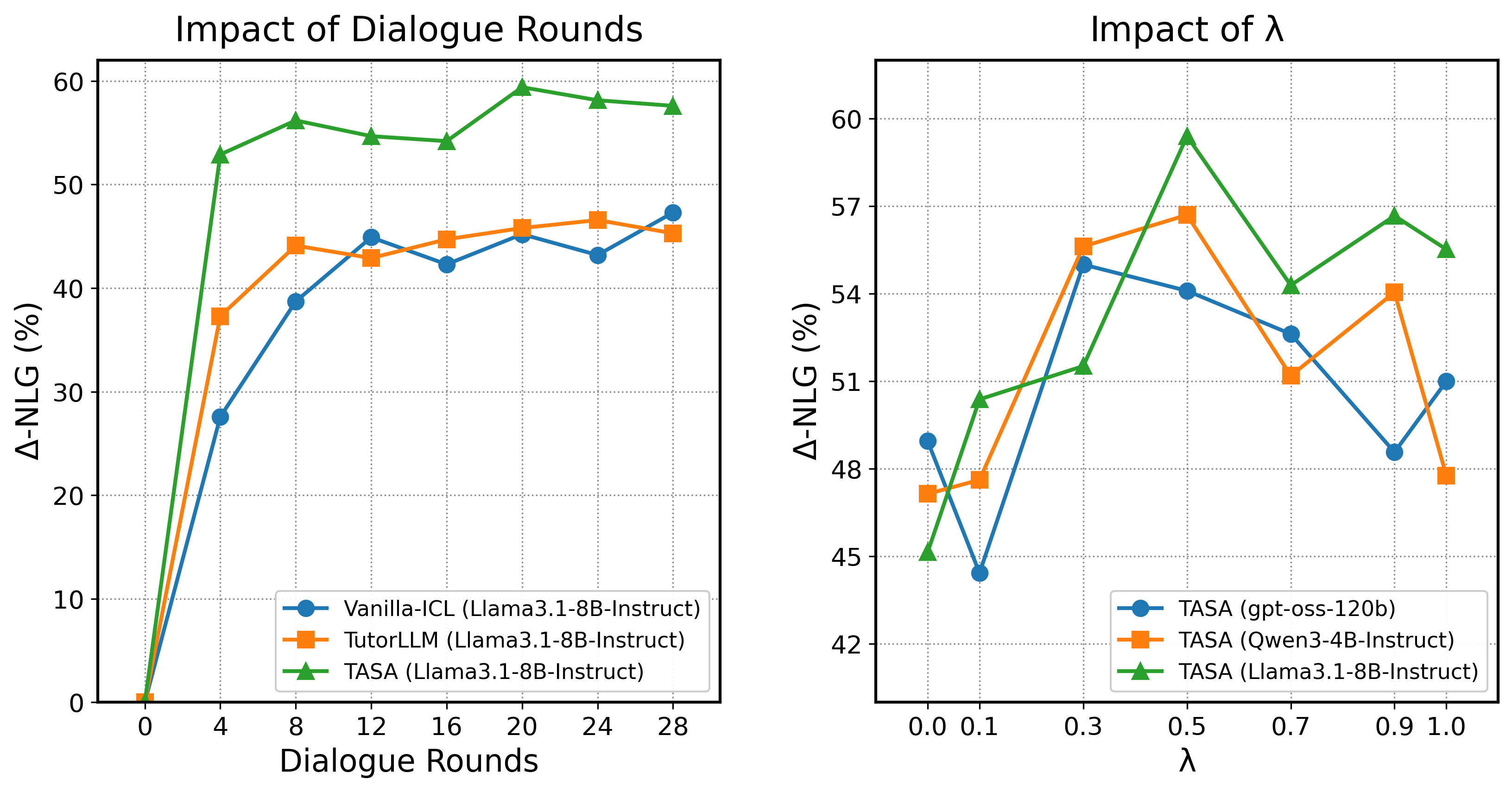}
\caption{Impact of dialogue rounds (left) and retrieval weight $\lambda$ (right) on normalized learning gain.}
\label{fig:ablation-round_lambda}
\end{figure}
To comprehensively evaluate TASA, we conduct four complementary studies examining: 1) module importance, 2) impact of dialogue rounds, 3) effect of the retrieval hyperparameter $\lambda$, and 4) comparison of different knowledge tracing (KT) backbones. All ablation experiments are performed on the Assist2017 dataset unless otherwise specified.

\subsubsection{Module Importance}
Table~\ref{tab:ablation_module} presents results when removing one key component at a time: persona (w/o Persona), memory (w/o Memory), or the forgetting score (w/o Forgetting). Removing any module leads to substantial degradation in both $\Delta$-NLG and Win Rate across all datasets. Specifically, eliminating the forgetting score results in an average drop of 6.8\% in $\Delta$-NLG, confirming that temporal retention modeling is crucial for adaptive instruction. Similarly, removing persona or memory decreases performance by 11.5\% and 4.9\% on average, respectively. Together, these results validate that all three components—persona, memory, and forgetting dynamics—are essential for effective personalized tutoring.

\subsubsection{Impact of Dialogue Rounds}
Figure~\ref{fig:ablation-round_lambda} (left) illustrates how $\Delta$-NLG evolves with the number of dialogue turns. Performance increases steadily from 4 to 20 rounds and gradually saturates thereafter, indicating that multi-turn interaction is essential for achieving stable learning improvement. Students benefit from iterative feedback and corrective practice across multiple turns, with diminishing returns beyond 20 rounds. This observation motivates the design of our default 20-round tutoring sessions. Across all dialogue lengths, TASA consistently outperforms the baseline methods, demonstrating its robustness and effectiveness under varying interaction frequencies. Moreover, the relative gains remain stable as the number of rounds increases, showing that TASA scales well with extended interactions. 

\subsubsection{Impact of Retrieval Weight $\boldsymbol{\lambda}$}

Figure~\ref{fig:ablation-round_lambda} (right) shows the effect of $\lambda$, which balances semantic similarity and concept-level alignment during persona and memory retrieval (Equation~\ref{eq:lambda}). TASA maintains stable performance within a broad range [0.3, 0.7] across all three backbones, with optimal results at $\lambda = 0.5$. The consistent trend across different $\lambda$ values indicates that the retrieval weighting mechanism can effectively balance semantic and conceptual relevance without requiring delicate parameter adjustments.

\begin{table}[t]
  \centering
  \renewcommand{\arraystretch}{1.15}
  \resizebox{\linewidth}{!}{%
    \begin{tabular}{lccccc}
      \toprule
      Method & Assist2017 & NIPS34 & Algebra2005 & Bridge2006 & Avg. \\
      \midrule
      \textbf{TASA\textsubscript{History}} 
        & 46.6~\std{0.6} 
        & 34.1~\std{0.7} 
        & 57.5~\std{0.9} 
        & 48.1~\std{0.6} 
        & 46.6 \\
      \textbf{TASA\textsubscript{AKT}} 
        & \underline{59.2~\std{0.9}} 
        & 43.1~\std{0.6} 
        & \underline{60.6~\std{1.4}} 
        & 39.7~\std{1.6} 
        & 50.7 \\
      \textbf{TASA\textsubscript{SimKT}} 
        & 56.1~\std{1.6} 
        & 44.4~\std{0.9} 
        & 47.3~\std{0.9} 
        & 25.1~\std{1.1} 
        & 43.2 \\
      \textbf{TASA\textsubscript{LPKT}} 
        & 52.5~\std{1.5} 
        & \underline{51.7~\std{1.3}} 
        & 55.4~\std{1.4} 
        & \textbf{54.3~\std{0.8}} 
        & \underline{53.5} \\
      \textbf{TASA\textsubscript{DKT}} 
        & \textbf{59.4~\std{1.0}} 
        & \textbf{54.9~\std{1.2}} 
        & \textbf{62.3~\std{1.5}} 
        & \underline{53.9~\std{1.7}} 
        & \textbf{57.6} \\
      \bottomrule
      \end{tabular}
      }
  \caption{Performance of TASA with different knowledge tracing models on the Llama3.1-8B-Instruct backbone.}

  \label{tab:kt_functions}
\end{table}

\subsubsection{Effect of Different KT Backbones}
Table~\ref{tab:kt_functions} compares TASA instantiated with different KT models: History-based mastery approximation (using historical correctness rates), AKT~\citep{ghosh2020context}, SimKT~\citep{liu2023simplekt}, LPKT~\citep{shen2021learning}, and DKT~\citep{piech2015deep}. While all variants achieve reasonable results, \textbf{TASA\textsubscript{DKT}} attains the highest average $\Delta$-NLG, outperforming the simple history-based approach by 11.0\% and other neural KT models by 9\% on average. LPKT also shows strong performance at 53.5\%, confirming that KT models designed to capture learning processes align well with our forgetting-aware framework. This demonstrates that accurate mastery estimation from the KT module directly enhances the effectiveness of temporal forgetting modeling and downstream tutoring quality.

\section{Conclusion \& Future Work}

We present TASA, a personalized tutoring framework that integrates persona, memory, and forgetting-aware mechanisms to enhance student mathematics learning. A temporal forgetting score decays students' persona and memory representations to capture the natural dynamics of learning. Empirical experiments and ablation studies across four mathematics knowledge-tracing benchmarks show that TASA consistently outperforms existing intelligent tutoring methods in both learning gain and tutoring personalization. In future work, we plan to extend TASA beyond mathematics to other educational domains and learner levels, incorporate multi-agent collaboration for richer interactions, and explore real-world deployment with human learners. We believe TASA provides a promising step toward cognitively grounded and adaptive LLM-based tutoring systems.

\clearpage



\bibliography{aaai2026}

\appendix

\clearpage
\section{Appendix A: Data Statistics}


In this section, we provide detailed statistics for the four mathematics tutoring benchmarks used in our evaluation. As shown in Table~\ref{tab:dataset_stats_rotated}, Assist2017 contains 1,708 students, 3,162 questions, 102 KCs, and 942,816 interactions. NIPS34 includes 4,918 students, 948 questions, 57 KCs, and 1,382,727 interactions. Algebra2005 comprises 574 students, 210,710 questions, 112 KCs, and 809,694 interactions, while Bridge2006 contains 1,138 students, 207,856 questions, 493 KCs, and 3,679,199 interactions. The notably higher question counts in Algebra2005 and Bridge2006 result from their step-level problem decomposition, where multi-step problems are broken into individual question instances. All datasets provide question IDs, knowledge concept annotations, and correctness labels, with Assist2017 additionally including response duration. These benchmarks collectively span a wide range of grade levels, question formats, and domain complexities, offering complementary perspectives for evaluating adaptive instruction. In particular, NIPS34 emphasizes short, knowledge-focused exercises with sparse interactions, while Bridge2006 represents large-scale, longitudinal student traces that challenge temporal generalization. Such diversity allows TASA to be rigorously tested under both micro-level learning diagnostics and long-term retention scenarios. Overall, these datasets provide a balanced and representative testbed for examining how well the proposed framework generalizes across heterogeneous educational settings.

\begin{table}[ht]
\centering
\renewcommand{\arraystretch}{1.2}
\resizebox{\linewidth}{!}{
\begin{tabular}{lcccc}
\hline\hline
\textbf{Dataset} & \textbf{Assist2017} & \textbf{NIPS34} & \textbf{Algebra2005} & \textbf{Bridge2006} \\
\hdashline
\multicolumn{5}{c}{\textit{Dataset Statistics}} \\
\hdashline
\# Students      & 1{,}708 & 4{,}918 & 574      & 1{,}138 \\
\# Questions     & 3{,}162 & 948     & 210{,}710 & 207{,}856 \\
\# KCs           & 102     & 57      & 112      & 493 \\
\# Interactions  & 942{,}816 & 1{,}382{,}727 & 809{,}694 & 3{,}679{,}199 \\
\hdashline
\multicolumn{5}{c}{\textit{Available Fields}} \\
\hdashline
Question ID          & \cmark & \cmark & \cmark & \cmark \\
Skill ID             & \cmark & \cmark & \cmark & \cmark \\
Results    & \cmark & \cmark & \cmark & \cmark \\
Duration   & \cmark & \xmark & \xmark & \xmark \\
\hline
\end{tabular}
}
\caption{Data statistics and available fields in four mathematics tutoring benchmarks.}
\label{tab:dataset_stats_rotated}
\end{table}

\section{Appendix B: Deriving and Analyzing the Forgetting Score}
\label{app:forgetting-derivation}

\subsection{Derivation}
Let $R_c(t)\in[0,1]$ be the retention for concept $c$ with $R_c(t)\big|_{\Delta t_c=0}=s_{t,c}$, and define the forgetting score $F_c(t)=1-R_c(t)$.
Assuming exponential decay of retention,
\begin{equation}
\begin{aligned}
\frac{\mathrm{d}}{\mathrm{d}(\Delta t_c)} R_c(t) &= -\frac{1}{S_c}\,R_c(t), \qquad S_c>0,\\
R_c(t) &= s_{t,c}\,\exp\!\Bigl(-\tfrac{\Delta t_c}{S_c}\Bigr),\\
F_c(t) &= 1 - s_{t,c}\,\exp\!\Bigl(-\tfrac{\Delta t_c}{S_c}\Bigr).
\end{aligned}
\end{equation}
To obtain a closed-form, low-cost surrogate, apply the Pad\'e-style approximation $\exp(-x)\approx \tfrac{1}{1+x}$ with $x=\Delta t_c/\tau$:
\begin{equation}
\begin{aligned}
R_c(t)
&= s_{t,c}\,\exp(-x)\\
&\approx s_{t,c}\,\frac{1}{1+x}\\
&= s_{t,c}\!\left(1-\frac{x}{1+x}\right)\\
&= s_{t,c} - s_{t,c}\,\frac{x}{1+x}\\
&= 1 - \bigl(1-s_{t,c}\bigr)\,\frac{x}{1+x}\\
&\approx 1 - \bigl(1-s_{t,c}\bigr)\,\tfrac{\Delta t_c}{\Delta t_c+\tau},
\end{aligned}
\end{equation}
hence
\begin{equation}
\begin{aligned}
F_c(t)
&= 1 - R_c(t)\\
&= 1 - s_{t,c}\,\exp\!\Bigl(-\tfrac{\Delta t_c}{S_c}\Bigr)\\
&\approx (1 - s_{t,c})\,\frac{\Delta t_c}{\Delta t_c+\tau}.
\end{aligned}
\end{equation}

\subsection{Analysis}
The rational surrogate preserves key qualitative properties while being numerically convenient:
\begin{equation}
\begin{aligned}
 0 \le &F_c(t) \le 1,\\
\frac{\partial F_c}{\partial \Delta t_c}
&= (1-s_{t,c})\,\frac{\tau}{(\Delta t_c+\tau)^2} \ge 0,\\
\frac{\partial F_c}{\partial s_{t,c}}
&= -\frac{\Delta t_c}{\Delta t_c+\tau} \le 0.
\end{aligned}
\label{equation:mono}
\end{equation}
\subsubsection{Monotonicity} Based on Equation~\ref{equation:mono}, $F_c(t)$ increases with elapsed time $\Delta t_c$ and decreases with mastery $s_{t,c}$.\\
\subsubsection{Boundary and Asymptotics} $F_c(0)=0$ and $F_c(t)\to 1-s_{t,c}$ as $\Delta t_c\to\infty$, yielding higher asymptotic forgetting for lower mastery.\\ 
\subsubsection{Approximation Accuracy} since $\exp(-x)=1-x+\tfrac{x^2}{2}-\cdots$ and $(1+x)^{-1}=1-x+x^2-\cdots$, the local error is $O(x^2)$ with leading term $\approx -\tfrac{1}{2}x^2$; thus the surrogate is tight near $\Delta t_c=0$ and remains stable for coarse time steps.\\
\subsubsection{Calibration} $S_c$ controls decay in the exponential form, while $\tau>0$ plays an analogous role in the surrogate; a practical choice is to select $\tau$ on validation data or tie it to concept-specific spacing intervals so that typical $\Delta t_c/\tau$ falls in a regime where the surrogate tracks observed forgetting.

\section{Appendix C: Prompt Details}
\label{app:Prompt}
In this section, we present the detailed evaluation protocol using large language models as impartial judges to assess the mathematics tutoring personalization and effectiveness of TASA and baseline methods. We also include the personalized tutoring generation prompt used in our proposed method.

\begin{prompt}{LLM as Judge Evaluation Prompt}
\textless\textbar im\_start\textbar\textgreater system\\
You are an expert educational AI evaluator. Your task is to determine which tutoring dialogue provides a more personalized and effective learning experience for the given student, based on their learning profile and history.

When comparing the two dialogues, consider the following aspects:

1. Personalization Depth – How well does the tutor adapt to the student’s known strengths, weaknesses, learning style, or past mistakes?

2. Contextual Relevance – Does the tutor refer appropriately to the student's previous learning records or specific problem areas?

3. Instructional Appropriateness – Are the explanations, feedback, and examples suitable for the student’s current level and needs?

4. Engagement and Responsiveness – Does the tutor respond dynamically to the student’s behavior, questions, or misunderstandings?

Important:  
- Your primary criterion is personalization quality — how well the dialogue tailors its teaching to this specific student.  
- However, reasonable instructional soundness (accuracy, clarity, pedagogical coherence) should also be considered when it directly affects personalization quality. 

Output your final judgment as:  
"Dialogue A is more personalized" or "Dialogue B is more personalized",  
followed by a short explanation (2--3 sentences) summarizing your reasoning.\\
\textless\textbar im\_end\textbar\textgreater\\

\textless\textbar im\_start\textbar\textgreater user\\
Two tutoring dialogues are provided, along with the student’s profile and learning history.  
Please evaluate which dialogue is more personalized and effective based on the above criteria.\\
\textless\textbar im\_end\textbar\textgreater
\end{prompt}

\begin{prompt}{Personalized Tutoring Prompt}
\textless\textbar im\_start\textbar\textgreater system\\
You are a personalized math tutor. Please generate the next instructional content that first explains the student's most recent response and then provides the next practice question, calibrated to the current retention state.\\ \textless\textbar im\_end\textbar\textgreater\\

\textless\textbar im\_start\textbar\textgreater user\\
Student Profile (Forgetting-Adjusted)\\
- [$\tilde{\text{p}}_1$, $\tilde{\text{p}}_2$, $\tilde{\text{p}}_3$]

Recent Learning Events (Forgetting-Adjusted)\\
- [$\tilde{\text{m}}_1$, $\tilde{\text{m}}_2$, $\tilde{\text{m}}_3$]

Current Dialogue Context\\
- [conversation history from $\mathcal{H}$]

Task:
Given the student's profile, recent events, and dialogue context, please generate instructional content that includes (1) a concise explanation of the student's last answer and (2) the next question tailored to the student's current knowledge state.  \\
\textless\textbar im\_end\textbar\textgreater
\end{prompt}



\end{document}